\documentclass[runningheads,a4paper]{llncs}

\usepackage{amsmath,amssymb}
\usepackage{mathtools}
\usepackage{xcolor}
\usepackage{xspace}
\usepackage{graphicx}
\usepackage{placeins}
\usepackage[basic,bold]{complexity}
\usepackage{tikz}
\usepackage{nicefrac}
\usepackage{hyperref}

\newcommand{\dataset}{\ensuremath{X}\xspace}
\newcommand{\datapoint}{\ensuremath{\vec{x}}\xspace}
\newcommand{\classifier}{\ensuremath{c}\xspace}
\newcommand{\trueclassifier}{\classifier_t\xspace}
\newcommand{\classifiers}{\ensuremath{C}\xspace}
\newcommand{\allClassifiers}{\ensuremath{\mathbb{C}}\xspace}

\newcommand{\labels}{\ensuremath{\ell}\xspace}
\newcommand{\labelset}{\ensuremath{[\ell]}\xspace}
\newcommand{\datasetLabelled}{\ensuremath{\mathcal{X}}\xspace}
\newcommand{\attack}{\ensuremath{{\delta}}\xspace}
\newcommand{\attackSet}{\ensuremath{A}\xspace}
\newcommand{\allAttacks}{\ensuremath{\Delta}\xspace}

\newcommand{\probDistr}{\ensuremath{\mathbb{P}}\xspace}
\newcommand{\prob}{\ensuremath{p}\xspace}
\newcommand{\lossZeroOne}{\ensuremath{\ell_{\mathrm{0-1}}}\xspace}
\newcommand{\loss}{\ensuremath{\ell}\xspace}

\newcommand{\norm}[1]{\left\lVert#1\right\rVert}

\newcommand{\wrt}{w.r.t.\@\xspace}

\newcommand{\Val}{\mathbf{Val}}

\usepackage{array,booktabs}
\newcolumntype{C}{>{$\displaystyle}c<{$}}
\usepackage[normalem]{ulem}
\useunder{\uline}{\ul}{}

\DeclareMathOperator*{\argmax}{arg\,max}

\DeclareMathOperator*{\expectop}{\mathbb{E}}
\newcommand{\Reals}{\ensuremath{\mathbb{R}}\xspace}    
\newcommand{\Naturals}{\ensuremath{\mathbb{N}}}
\newcommand{\Ireal}{[0,\, 1]\subseteq\mathbb{R}}  
\newcommand{\Ex}{\ensuremath{\expectop}\xspace}        

\newcommand{\Distr}{\mathit{Distr}}

\newcommand{\distDom}{D}

\newcommand{\distFunc}{\mu}
\newcommand{\distDomElem}{x}

\makeatletter
\let\vec\mathbf
\let\epsilon\varepsilon
\let\phi\varphi

\makeatother

\begin{document}

\title{Robustness Verification for Classifier Ensembles%
\thanks{Research funded by the project NWA.BIGDATA.2019.002: ``EXoDuS - EXplainable Data Science'' and the
FWO G030020N project ``SAILor''}}

\author{Dennis Gross\inst{1}
\and Nils Jansen\inst{1}
\and Guillermo A. P\'erez\inst{2}
\and Stephan Raaijmakers\inst{3}
}

\authorrunning{D. Gross et al.}

\institute{Radboud University Nijmegen, The Netherlands
\and University of Antwerp, Belgium
\and TNO \& Leiden University, The Netherlands
}
\maketitle

\begin{abstract}
We give a formal verification procedure that decides whether a classifier ensemble is robust against arbitrary randomized attacks.
Such attacks consist of a set of deterministic attacks and a distribution over this set. 
The robustness-checking problem consists of assessing, given a set of classifiers and a labelled data set, whether there exists a randomized attack that induces a certain expected loss against all classifiers.
We show the NP-hardness of the problem and provide an upper bound on the number of attacks that is sufficient to form an optimal randomized attack.
These results provide an effective way to reason about the robustness of a classifier ensemble. 
We provide SMT and MILP encodings to compute optimal randomized attacks or prove that there is no attack inducing a certain expected loss. 
In the latter case, the classifier ensemble is provably robust.
Our prototype implementation verifies multiple neural-network ensembles trained for image-classification tasks.
The experimental results using the MILP encoding are promising both in terms of scalability and the general applicability of our verification procedure. 
\end{abstract}

\section{Introduction}
Recent years have seen a rapid progress in \emph{machine learning} (ML) with a strong impact to fields like autonomous systems, computer vision, or robotics.
As a consequence, many systems employing ML show an increasing interaction with aspects of our everyday life, consider autonomous cars operating amongst pedestrians and bicycles.
While studies indicate that self-driving cars, inherently relying on ML techniques, make around 80\% fewer traffic mistakes than human drivers~\cite{nyholm2018the},  verifiable \emph{safety} remains a major open challenge~\cite{stoica2017berkeley,freedman2016safety,future_AI,amodei2016concrete}.

In the context of self-driving cars, for instance, certain camera data may contain noise that can be introduced randomly or actively via so-called adversarial attacks.
We focus on the particular problem of such attacks in image classification.
A successful attack perturbs the original image in a way such that a human does not recognize any difference while ML \emph{misclassifies} the image.
A measure of difference between the ground truth classification, for instance by a human, and a potentially perturbed ML classifier is referred as the \emph{loss}.

A standard way to render image classification more robust against adversarial attacks is to employ a set of classifiers, also referred to as \textit{classifier ensembles}~\cite{DBLP:conf/iclr/AbbasiG17,DBLP:journals/corr/abs-1906-02816,pinot2020randomization,abbasi2020toward}. 
The underlying idea is to obscure the actual classifier from the attacker.
One possible formalization of the competition between an adversarial attacker and the ensemble is that of a
zero-sum game~\cite{DBLP:journals/corr/abs-1906-02816}:
The attacker chooses first, the ensemble tries to react to the attack with minimal loss --- that is, choosing a classifier that induces maximal classification accuracy.

\begin{figure}
    \centering
      \begin{tikzpicture}[yscale=0.5]
    \foreach \i in {0,...,5}
        \draw 
            (\i,2.5pt) -- +(0,-5pt) node [anchor=north, font=\small] {$\i$}
        ;
    \draw [->] (0,0) -- (6,0) node [anchor=west] {$x$};
    \draw [->] (0,0) -- (0,2) node [anchor=south] {$y$};
    \fill [green,opacity=0.3] (4,0) rectangle (6,2);
    \fill [blue,opacity=0.3] (0,0) rectangle (2,2);
    \draw [->,ultra thick,blue] (2, 0) -- (2, 2);
    \draw [->,ultra thick,dashed,green] (4,0) -- (4, 2);
    \draw [->,red,densely dotted] (3,1) -- (1,1);
    \draw [->,red] (3,1) -- (5,1);
    \filldraw [black] (3,1) circle (2pt);
  \end{tikzpicture}
    \caption{We depict a single data point in $\mathbb{R}^2$ with label $1$. The region to the left of the solid vertical line
    corresponds to points labelled with $2$ by one classifier; the region to the right of the dashed line, those labelled with $2$
    by another classifier. Both (linear) classifiers label all other points in $\mathbb{R}^2$ with $1$. (Hence, they correctly label the data point with $1$.)
    The dotted attack, moving the data point left, induces a misclassification of the point by one of the classifiers. The solid
    attack, moving the data point right, induces a misclassification of the point by one of the classifiers. Note that every attack has a classifier which is ``robust'' to it, i.e. it does not misclassify the perturbed point. However,
    if the attacker chooses an attack uniformly at random, both of them misclassify the point with probability $\nicefrac{1}{2}$.}
    \label{fig:why-rand}
\end{figure}
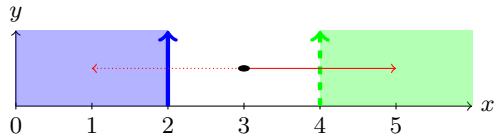

In this setting, the attacker may need to use randomization to behave optimally (see Fig.~\ref{fig:why-rand}, cf.~\cite{ag11}).
Such an attack is called optimal if the \emph{expected loss} 
is maximized regardless of the choice of classifier.


Inspired by previous approaches for single classifiers~\cite{DBLP:conf/cav/KatzBDJK17,DBLP:conf/atva/Ehlers17}, we develop a formal verification procedure that decides if a classifier ensemble is \textit{robust} against any randomized attack.
In particular, the formal problem is the following.
Given a set of classifiers and a labelled data set, we want to find a probability distribution and a set of attacks that induce an optimal randomized attack. 
 Akin to the setting in~\cite{DBLP:journals/corr/abs-1906-02816}, one can provide thresholds on potential perturbations of data points and the minimum shift in classification values.
Thereby, it may happen that no optimal attack exists, in which case we call the classifier ensemble \emph{robust}.
Our aim is the development of a principled and effective method that is able to either find the optimal attack or prove that the ensemble is robust with respect to the predefined thresholds. 

\begin{figure}[h]
\centering
\begin{tikzpicture}[->,yscale=0.6,el/.style={font=\scriptsize},inner sep=1pt]
    \draw (3,0) rectangle (7,4);
    \node[font=\Large] (v) at (5,2) {Verifier};
    
    \path (7,2) edge node[el,above]{ROBUST/NOT ROBUST}
                     node[el,below]{Probabilities \& attacks} (11,2);
    \path (0,3.3) edge node[el,above]{Set of classifiers}  (3,3.3);
    \path (0,2.6) edge node[el,above]{Set of data points}  (3,2.6);
    \path (0,1.9) edge node[el,above]{Number of attacks}   (3,1.9);
    \path (0,1.2) edge node[el,above]{Max attack strength} (3,1.2);
    \path (0,0.5) edge node[el,above]{Min prediction}
                       node[el,below]{perturbation} (3,0.5);
    
\end{tikzpicture}
\caption{The verifier takes as input a set of classifiers, a set of labelled data points, the number of attacks, and the attack properties. If the verifier does not find a solution, we can be sure is robust against any attack with the specific properties. Otherwise, it returns the optimal attack.}
\label{fig:verifier}
\end{figure}
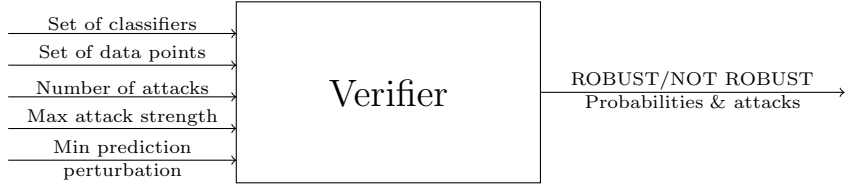

To that end, we first establish a number of theoretical results.
First, we show that the underlying formal problem is \NP-hard.
Towards computational tractability, we also show that for an optimal attack there exists an upper bound on the number of attacks that are needed. 
Using these results, we provide an SMT encoding that computes suitable randomized attacks for a set of convolutional neural networks with ReLU activation functions and a labelled data set.
In case there is no solution to that problem, the set of neural networks forms a robust classifier ensemble, see Fig.~\ref{fig:verifier}.
Together with the state-of-the-art SMT solver Z3~\cite{de2008z3}, this encoding provides a complete method to solve the problem at hand.
Yet, our experiments reveal that it scales only for small examples.
We outline the necessary steps to formulate the problem as a mixed-integer linear programming (MILP), enabling the use of efficient optimization solvers like Gurobi~\cite{gurobi}.

In our experiments, we show the applicability of our approach by means of a benchmark set of binary classifiers, which were trained on the MNIST and German traffic sign datasets~\cite{deng2012mnist,stallkamp2011german}.

\subsection*{Related work}
It is noteworthy that there is some recent work on
robustness checking of decision-tree ensembles~\cite{ranzatorobustness}. However, their approach
is based on abstract interpretation and is thus not complete.
Other approaches for robustness checking of machine learning classifiers focus on single classifiers (see, {e.g.,~\cite{DBLP:conf/cav/KatzBDJK17,DBLP:conf/atva/Ehlers17,DBLP:conf/nips/BunelTTKM18,DBLP:conf/concur/Kwiatkowska19,popl-vechev})}. 
Akin to our approach, {some of these works} employ SMT solving.
In~\cite{DBLP:conf/aaai/AkintundeKLP19}, MILP-solving is used for verification tasks on (single) recurrent neural networks.
In contrast, our framework allows to compute attacks for classifier ensembles.

In~\cite{sanjit-robust}, Dreossi et al.
propose a robustness framework
which unifies the optimization and verification views on the
robustness-checking problem
and encompasses several existing approaches.
They explicitly mention that their framework applies to \emph{local
robustness} and argue most of the existing work on finding
adversarial examples and verifying robustness fits their framework.
Our work, when we have a single classifier and a singleton data set,
fits precisely into their framework. However, we
generalize in those two dimensions by averaging over the
\emph{robustness target value} (in their jargon) for all points in a
data set, and by considering ensemble classifiers.
This means that our point of view of the \emph{adversarial
environment} is neither that of a white-box attacker nor is it a
black-box attacker. Indeed, we know the set of classifiers but we do
not know what strategy is used to choose which (convex combination of)
classifiers to apply. Our environment is thus a gray-box
attacker.
\section{Preliminaries}
Let $\datapoint$ be a vector $(x_1,\dots, x_d) \in \mathbb{R}^d$. 
We write $\norm{\datapoint}_1$ for the ``Manhattan norm'' of $\datapoint$, that is $\sum_{i=1}^d |\datapoint_i|$.

We will make use of a partial inverse of the $\max$ function. 
Consider a totally ordered set $Y$ and a function $f\colon X \to Y$. 
Throughout this work we define the $\argmax$ (arguments of the maxima) partial function as follows.
For all $S \subseteq X$ we set $\argmax_{s \in S} f(s) := m$ if $m$ is the \emph{unique} element of $S$ such that $f(m) = \max_{s \in S} f(s)$. 
If more than one element of $S$ witnesses the maximum then $\argmax_{s \in S} f(s)$ is \emph{undefined}.

A \emph{probability distribution} over a finite set $\distDom$ is a function
$\distFunc\colon\distDom\rightarrow\Ireal$ with
$\sum_{\distDomElem\in\distDom}\distFunc(\distDomElem)=1$.  The set of all
distributions on $\distDom$ is $\Distr(\distDom)$.

\subsection{Neural networks}
We loosely follow the neural-network notation
from~\cite{DBLP:conf/cav/KatzBDJK17,DBLP:conf/atva/Ehlers17}.
A \emph{feed-forward neural network} (NN for short) with $d$ inputs and $\labels$
outputs encodes a function $f\colon \mathbb{R}^d \to \mathbb{R}^\labels$. 
We focus on NNs with \emph{ReLU} activation functions. 
Formally, the function $f$ is given in the form of
\begin{itemize}
  \item a sequence $\vec{W}^{(1)},\dots,\vec{W}^{(k)}$ of \emph{weight matrices} with
    $\vec{W}^{(i)} \in \mathbb{R}^{d_i \times d_{i-1}}$, for all $i = 1,\dots,k$,
    and
  \item a sequence $\vec{B}^{(1)},\dots,\vec{B}^{(k)}$ of \emph{bias vectors} with
    $\vec{B}^{(i)} \in \mathbb{R}^{d_i}$, for all $i = 1, \dots,k$. 
\end{itemize}
Additionally, we have that {$d_0,\ldots,d_k\in\Naturals$ with} $d_0 = d$ and $d_k
= \labels$. 
We then set $f = g^{(k)}(\datapoint)$ for all $\datapoint \in \mathbb{R}^d$ where
for all $i = 1,\dots,k$ we define
\[
  g^{(i)}(\datapoint) := \mathrm{ReLU}(\vec{W}^{(i)} g^{(i-1)}(\datapoint) + \vec{B}^{(i)}),
\]
and $g^{(0)}(\datapoint) := \datapoint$. 
The ReLU function on vectors $\vec{u}$ is the element-wise maximum between $0$ and the vector entries, that is, if $\vec{v} = \mathrm{ReLU}(\vec{u})$ then $\vec{v}_{i} = \max(0,\vec{u}_{i})$.

We sometimes refer to each $g^{(i)}$ as a \emph{layer}. Note that each layer is
fully determined by its corresponding weight and bias matrices.

\subsection{Neural-network classifiers}
A \emph{data set} $\dataset\subseteq\Reals^d$ is a finite set of (real-valued) data
points $\datapoint\in\Reals^d$ of dimension $d\in\Naturals_{>0}$.
A \emph{classifier} $\classifier\colon\dataset\rightarrow \labelset$ is a
partial function that attaches to each data point a label from $\labelset =
\{1,\dots,\labels\}$, the \emph{set of labels}. 
We denote the set of all classifiers over $\dataset$ by $\allClassifiers$. 
An NN-encoded classifier is simply a partial function $f\colon \mathbb{R}^d
\to \mathbb{R}^k$ given as an NN that assigns to each data point $\datapoint
\in \mathbb{R}^d$ the label $\argmax_{i \in \labelset} h(i)$ where $h(i) =
f(\datapoint)_i$.
Intuitively, the label is the index of the largest entry in the vector resulting from applying $f$ to $\datapoint$. 
Note that if the image of $x$ according to $f$ has several maximal entries then the $\argmax$ and the output label are undefined.

\begin{definition}[Labelled data set]
A \emph{labelled data set} $\datasetLabelled=(\dataset,\trueclassifier)$ consists of a data set $\dataset$ and a total classifier $\trueclassifier$ for $\dataset$, i.e. $\trueclassifier$ is a total function.
\end{definition}
In particular, $\trueclassifier(\datapoint)$ is defined for all $\datapoint \in \dataset$ and considered to be the ``ground truth'' classification for the whole data set $\dataset$.

\section{Problem Statement}
We state the formal problem and provide the required notation.
Recall that in our setting we assume to have an ensemble of classifiers $\classifiers\subseteq\allClassifiers$.
Such an ensemble is attacked by a set of attacks that are selected randomly.

\begin{definition}[Deterministic attack]\label{def:det_attack}
A \emph{deterministic attack} for a labelled data set $(\dataset,\trueclassifier)$
and a classifier $\classifier\colon\dataset\rightarrow \labelset$
is a
function $\attack\colon\dataset\rightarrow\Reals^d$.
An attack $\attack$ induces a \emph{misclassification} for
$\datapoint\in\dataset$ and $\classifier$ if
$\classifier(\datapoint+\attack(\datapoint)) \neq \trueclassifier(\datapoint)$
or if $\classifier(\datapoint+\attack(\datapoint))$ is undefined.
The set of all deterministic attacks is $\allAttacks$.
An attack is $\epsilon$-bounded if $\norm{\attack(\datapoint)}_1\leq\epsilon$ holds for all $\datapoint \in X$.
\end{definition}
We sometimes call the value $\datapoint+\attack(\datapoint)$ the \emph{attack point}.
{Note that the classifier $\classifier$ is not perfect}, that is,
$\classifier(\datapoint)\neq\trueclassifier(\datapoint)$ for some $\datapoint\in\dataset$, already a zero-attack $\attack(\datapoint)=0$ leads to a misclassification.

We extend deterministic attacks by means of probabilities.
\begin{definition}[Randomized attack]\label{def:rand_attack}
A finite set $\attackSet\subseteq\allAttacks$ of deterministic attacks together with a probability
distribution $\probDistr \in\Distr(\attackSet)$ is a \emph{randomized attack}
$(\attackSet,\probDistr)$. 
A randomized attack is \emph{$\epsilon$-bounded} if for all attacks $\attack\in\attackSet$ with $\probDistr(\attack)>0$ it holds that $\norm{\attack(\datapoint)}_1\leq\epsilon$ for all $\datapoint \in \dataset$.
\end{definition}

\begin{figure}[ht]
    \centering
      \begin{tikzpicture}[xscale=0.8,yscale=0.7]
    \foreach \i in {0,...,5}
        \draw 
            (\i,2.5pt) -- +(0,-5pt) node [anchor=north, font=\small] {$\i$}
        ;
    \foreach \i in {0,...,5}
        \draw
            (2.5pt,\i) -- +(-5pt,0) node [anchor=east, font=\small] {$\i$}
        ;
    \draw [->] (0,0) -- (6,0) node [anchor=west] {$x$};
    \draw [->] (0,0) -- (0,6) node [anchor=south] {$y$};
    \fill [green,opacity=0.3] (4,0) rectangle (6,6);
    \fill [blue,opacity=0.3] (0,0) rectangle (2,6);
    \draw [->,ultra thick,blue] (2, 0) -- (2, 6);
    \draw [->,ultra thick,dashed,green] (4,0) -- (4, 6);
    \draw [red, densely dotted] (1,3) -- (3,5) -- (5,3) -- (3,1) -- (1,3);
    \node [red,font=\scriptsize,anchor=north,align=center] at (3,1) {$2$-bounded\\ attacks}; 
    \draw [->,red] (3,3) -- node[above,pos=0.3,font=\scriptsize]{$\delta_1$} (1,3);
    \draw [->,red] (3,3) -- node[above,pos=0.3,font=\scriptsize]{$\delta_2$} (5,3);
    \filldraw [black] (3,3) circle (2pt);
  \end{tikzpicture}
    \caption{The dotted diamond contains the set of all $2$-bounded attacks in the setting described in Fig.~\ref{fig:why-rand}. Both $\delta_1$ and $\delta_2$
    are therefore $2$-bounded (deterministic) attacks. Hence, any randomized attack with $A = \{\delta_1,\delta_2\}$ is also $2$-bounded.}
    \label{fig:eps-bound}
\end{figure}
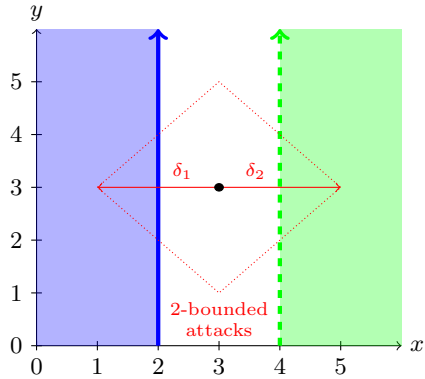

In general, a \emph{loss function} $\loss\colon\allClassifiers\times\Reals^d\times\Reals^d\rightarrow\Reals$ describes the \emph{penalty} incurred by a classifier with respect to a labelled data point and an attack. 
In this work, we will focus on the widely used zero-one loss.
\begin{definition}[Zero-one loss function]\label{def:zero_one_loss}
The \emph{$(0-1)$-loss function} $\lossZeroOne
\colon\allClassifiers\times\Reals^d\times\Reals^d\rightarrow\{0,1\}$ for
a labelled data set $(\dataset,\trueclassifier)$, a classifier
$\classifier\colon\dataset\rightarrow \labelset$,
and a deterministic attack $\attack\in\attackSet$ is given by the following
for all $\datapoint \in \dataset$
\begin{align*}
	\lossZeroOne(\classifier,\datapoint,\attack(\datapoint))=
	\begin{cases}
	  0 & \text{if } \classifier(\datapoint+\attack(\datapoint))=\trueclassifier(\datapoint)\\
	  1 & \text{otherwise.}	
	\end{cases}
\end{align*}
\end{definition}
In particular, the loss function yields one if the image of the classifier $\classifier$ is undefined for the attack point $\datapoint+\attack(\datapoint)$.
Note furthermore that the loss is measured with respect to the ground truth classifier $\trueclassifier$. 
Thereby, the classifier $\classifier$ and the zero function as
deterministic attack do not necessarily induce a
loss of zero with respect to $\trueclassifier$.
This assumption is realistic as, while we expect classifiers to perform well with regard to the ground truth, we cannot assume perfect classification in realistic settings.

We now connect a randomized attack to an ensemble, that is, a finite set $\classifiers\subseteq\allClassifiers$ of classifiers. 
In particular, we quantify the overall value a randomized attack induces with respect to the loss function and the ensemble.

\begin{definition}[Misclassification value]\label{def:misclassification}
The \emph{misclassification value} of a randomized attack $(\attackSet,\probDistr)$ with
respect to a labelled data set $\datasetLabelled = (\dataset,\trueclassifier)$ and a finite
set of classifiers $\classifiers\subseteq\allClassifiers$ is given by
\begin{equation}\label{eqn:robustness-value}
  \Val(\attackSet,\probDistr) :=
  \min_{\classifier\in\classifiers}\,
  \frac{1}{|\dataset|}\sum_{\datapoint\in\dataset}\,
  \Ex_{\attack\sim\probDistr}[\lossZeroOne(\classifier,\datapoint,\attack(\datapoint))].
\end{equation}
	
\end{definition}
This value is the minimum (over all classifiers) mean expected loss with respect to the randomized attack and the classifiers from $\classifiers$.
An \emph{optimal adversarial attack} against $\classifiers\subseteq\allClassifiers$ with respect to a labelled data set
$(\dataset,\trueclassifier)$ is a randomized attack which maximizes the value $\Val(\attackSet,\probDistr)$ in
Equation~\eqref{eqn:robustness-value}.

We are now ready to formalize a notion of robustness in terms of $\epsilon$-bounded attacks and a \emph{robustness bound} $\alpha\in\Reals$, as proposed in~\cite{DBLP:journals/corr/abs-1906-02816} for a set of classifiers.
\begin{definition}[Bounded robustness]\label{def:robustness}
A set of classifiers $\classifiers\in\allClassifiers$ for a labelled data set $(\dataset,\trueclassifier)$ is called \emph{robust bounded by $\epsilon$ and $\alpha$} ($\epsilon,\alpha$-robust)
if it holds that 
\begin{equation}\label{eqn:robustness-bound}
  \forall{(\attackSet,\probDistr)\in 2^{\allAttacks}\times\Distr(\attackSet)}.\,
  \Val(\attackSet,\probDistr) < \alpha,
\end{equation}
where the $(\attackSet,\probDistr$) range over all $\epsilon$-bounded randomized attacks.
\end{definition}

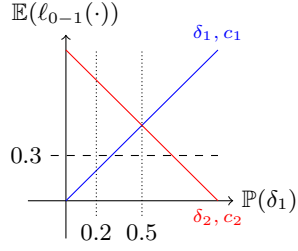
\begin{figure}[ht]
    \centering
    \begin{tikzpicture}
      \draw[->] (-0.5,0) -- (2.2,0) node[right] {$\probDistr(\delta_1)$};
      \draw[->] (0,-0.5) -- (0,2.2) node[above] {$\mathbb{E}(\ell_{0-1}(\cdot))$};
      \draw[scale=2,domain=0:1,smooth,variable=\x,blue] plot ({\x},{\x}) node[above,font=\scriptsize]{$\delta_1,c_1$};
      \draw[scale=2,domain=0:1,smooth,variable=\x,red]  plot ({\x},{1-\x}) node[below,font=\scriptsize]{$\delta_2,c_2$};
      \draw[dashed] (-0.2,0.6) node [anchor=east, font=\small] {$0.3$} -- (2,0.6);
      \draw[densely dotted] (0.4,-0.2) node [anchor=north, font=\small] {$0.2$}
      -- (0.4,2);
      \draw[densely dotted] (1,-0.2) node [anchor=north, font=\small] {$0.5$}
      -- (1,2);
\end{tikzpicture}
    \caption{Continuing with the example from Figs~\ref{fig:why-rand} and~\ref{fig:eps-bound}, we now plot the expected loss per attack
    and the corresponding classifier. (That is, the classifier which misclassifies
    the perturbed point.) On the horizontal axis we have the probability $x$ assigned to $\delta_1$ and we assume $\probDistr(\delta_2) = 1 - x$. Note
    that $x = 0.2$ is such that the minimal expected loss, i.e. the misclassification value, is strictly less
    than $0.3$. Indeed, one classifier manages to correctly classifier the perturbed point with probability $0.8$ in this case. With $x=0.5$ we see
    that the misclassification value is $0.5$. Hence, the ensemble is not
    $(2,0.5)$-robust.}
    \label{fig:alpha-bound}
\end{figure}

In other words, an $(\epsilon,\alpha)$-robust ensemble is such that for all possible $\epsilon$-bounded random
attacks $(\attackSet,\probDistr)$, there is at least one classifier $c\in\classifiers$ from the ensemble such that
\(
\sum_{\datapoint\in\dataset}\,
  \Ex_{\attack\sim\probDistr}[\lossZeroOne(\classifier,\datapoint,\attack(\datapoint))] < \alpha
  |X|.
\)
Conversely, an ensemble is not $(\epsilon,\alpha)$-robust if there is an $\epsilon$-bounded randomized attack with a misclassification value of at
least $\alpha$.

\section{Theoretical Results}
In this section we establish two key results that carry essential practical implication for our setting.
First, we show that in order to obtain an optimal randomized attack, only a bounded number of deterministic attacks is needed.\footnote{An analogue of this
property had already been observed by Perdomo and Singer in~\cite[Section
3]{DBLP:journals/corr/abs-1906-02816} in the case when classifiers are
chosen randomly.}
Thereby, we only need to take a bounded number of potential attacks into
account in order to prove the $\alpha$-robustness of a set of classifiers and a given labelled data set. 
Second, we establish that our problem is in fact \NP-hard, justifying the use
of SMT and MILP solvers to (1) compute any optimal randomized attack and, more importantly, to (2) prove robustness against any such attack.

\subsection{Bounding the number of attacks}\label{sec:bounded_number_of_attacks}
In the following, we assume a fixed labelled data set $(\dataset,\trueclassifier)$.
For every classifier $\classifier \in \classifiers$ and every deterministic attack $\attack\in\allAttacks$, let us write $M_{\classifier}(\attack)$ to denote the value $\sum_{x \in X}
\lossZeroOne(\classifier,\datapoint,\attack(\datapoint))$.
Observe that $0 \leq M_c(\attack) \leq |\dataset|$ for all $\classifier\in\classifiers$ and $\attack\in\allAttacks$.
Furthermore, for all $\classifier\in\classifiers$ and randomized attacks $(\attackSet,\probDistr)$ it holds that:
\begin{align*}
  & \sum_{\datapoint\in\dataset}\,
  \Ex_{\attack\sim\probDistr}[\lossZeroOne(\classifier,\datapoint,\attack(\datapoint))]
  =  \sum_{\datapoint\in\dataset} \sum_{\attack \in \attackSet} \probDistr(\attack)\cdot
  \lossZeroOne(\classifier,\datapoint,\attack(\datapoint))\\
  & = \sum_{\attack \in \attackSet} \probDistr(\attack) \underbrace{\left(\sum_{\datapoint \in \dataset}
  \lossZeroOne(\classifier,\datapoint,\attack(\datapoint))\right)}_{=M_{\classifier}(\attack)}
  = \sum_{\attack \in \attackSet} \probDistr(\attack) \cdot M_{\classifier}(\attack)
\end{align*}
We get that Equation~\eqref{eqn:robustness-bound} from Def.~\ref{def:misclassification} is false if and only
if the following holds.
\begin{equation}\label{eqn:simple}
  \exists{(\attackSet,\probDistr)\in 2^{\allAttacks}\times\Distr(\attackSet)}.\,
  \forall{\classifier \in \classifiers}.\, 
  \sum_{\attack \in \attackSet} \probDistr(\attack) \cdot M_{\classifier}(\attack) \geq \alpha|\dataset|
\end{equation}

\begin{proposition}[Bounded number of attacks]\label{prop:bnoa}
Let $\alpha \in \Reals$ and consider the labelled data set $(\dataset,\trueclassifier)$ together with a finite set of classifiers $\classifiers \subseteq \allClassifiers$. 
For all randomized attacks $(\attackSet,\probDistr)$, there exists a randomized attack $(\attackSet',\probDistr')$ such that
  \begin{itemize}
    \item $|\attackSet'| \leq (|\dataset|+1)^{|\classifiers|}$,
    \item $\Val(\attackSet',\probDistr') = \Val(\attackSet,\probDistr)$, and
    \item $(\attackSet',\probDistr')$ is $\epsilon$-bounded if
      $(\attackSet,\probDistr)$ is $\epsilon$-bounded.
  \end{itemize}
\end{proposition}
\begin{proof}
  We proceed by contradiction. 
  Let $(\attackSet,\probDistr)$ be an $\epsilon$-bounded randomized attack with a misclassification value of $\alpha$ such that $|\attackSet| > (|\dataset|+1)^{|\classifiers|}$. 
  Further suppose that $(\attackSet,\probDistr)$ is minimal (with respect to the size of $\attackSet$) amongst all such randomized attacks. 
  It follows that there are attacks $\attack,\attack' \in \attackSet$ such that $M_c(\attack) = M_c(\attack')$
  for all $\classifier \in \classifiers$. We thus have that
  \[
    \probDistr(\attack) \cdot M_c(\attack) +
    \probDistr(\attack') \cdot M_c(\attack')
    = \left( \probDistr(\attack) + \probDistr(\attack') \right) \cdot
    M_c(\attack).
  \]
  
  \noindent
  Consider now the randomized attack $(\attackSet',\probDistr')$ obtained by modifying
  $(\attackSet,\probDistr)$ so that $\probDistr(\attack) = \probDistr(\attack) +
  \probDistr(\attack')$ and $\attack'$ is removed from $\attackSet$. From the above
  discussion and Equation~\eqref{eqn:simple} it follows that
  $(\attackSet',\probDistr')$, just like $(\attackSet,\probDistr)$, has a misclassification value of
  $\alpha$. Furthermore, since $\attackSet' \subseteq A$, we have that
  $(\attackSet',\probDistr')$ is $\epsilon$-bounded and that $|\attackSet'| < |\attackSet|$. This
  contradicts our assumption regarding the minimality of
  $(\attackSet,\probDistr)$ amongst $\epsilon$-bounded randomized attacks with the same value $\alpha$.\qed
\end{proof}

\subsection{NP hardness of non-robustness checking}
It is known that checking whether linear properties hold for a given NN with
ReLU activation functions is \NP-hard~\cite{DBLP:conf/cav/KatzBDJK17}. We
restate this using our notation.
\begin{proposition}[From~{\cite[Appendix I]{DBLP:conf/cav/KatzBDJK17}}]
  \label{pro:reluplex}
  The following problem is \NP-hard: Given an NN-encoded function $f\colon
  \Reals^n \to \Reals^m$ and closed nonnegative intervals $(I_k)_1^n,(O_\ell)_1^m$, decide whether there exists $\vec{x} \in \prod_{k=1}^n I_k$ such that $f(\vec{x}) \in \prod_{\ell=1}^m O_\ell$.
\end{proposition}
Intuitively, determining whether there {exists} a point in a given \emph{box} ---
that is, a hypercube defined by a Cartesian product of intervals --- from
$\Reals^n$ whose image according to $f$ is in a given box from
$\Reals^m$ is \NP-hard.
We will now reduce this to the problem of determining if
there is a randomized attack such that its misclassification value takes at least
a given threshold.
\begin{theorem}\label{theorem:np}
  The following problem is \NP-hard: For a labelled data set
  $\datasetLabelled=(\dataset,\trueclassifier)$, a set $\classifiers$ of classifiers, and a value
  $\alpha \in \mathbb{Q}$, decide if there exists an $\epsilon$-bounded randomized attack $(\attackSet,\probDistr)$ \wrt $\datasetLabelled$ and $\classifiers$ such that $\Val(\attackSet,\probDistr) \geq \alpha$.
\end{theorem}
\begin{proof}
  We use Proposition~\ref{pro:reluplex} and show how to construct, for any
  NN-encoded function $g\colon \Reals^n \to \Reals$ and any constraint
  $\ell \leq g(\vec{x}) \leq u$, two classifiers $c_\ell,c_u$ such that a single deterministic attack $\delta$
  causes $\vec{0}$, the single data point,
  to be misclassified by
  both $c_\ell$ and $c_u$ if and only if the
  constraint holds. Note that the
  $\epsilon$ bound can be chosen to be large enough so that it contains the
  box $\prod_{k=1}^n I_k$ and that the identity function over nonnegative
  numbers is NN-encodable, that is, using the identity matrix as weight matrix
  $\vec{W}$ and a zero bias vector $\vec{B}$.
  For every instance of the problem from Proposition~\ref{pro:reluplex} we can
  therefore construct $2(n+m)$ NNs that encodes all input and output constraints:
  $2n$ of them based on the identity function to encode input constraints and
  $2m$ based on the input NN from the given instance. 
  It follows that determining if there exists an $\epsilon$-bounded deterministic attack
  $\attack$ that causes $\vec{0}$ to be \emph{simultaneously
  misclassified} by a given classifier ensemble is \NP-hard.
   Hence, to conclude, it suffices to argue that the latter problem reduces to our
  robustness-value threshold problem.
  The result follows from Lemmas~\ref{lem:intervals}
  and~\ref{lem:all-mis}.\qed
\end{proof}

\subsubsection{Enforcing interval constraints.}\label{sec:intervals}
Let $g \colon \Reals^n \to \Reals$ be an NN-encoded function and $\ell,u \in \mathbb{R}$ with $\ell \leq u$. Consider now
the constraint $\ell \leq g(\vec{x}) \leq u$. 
Henceforth we will focus on the labelled data set $(X,\trueclassifier)$ with
  $X = \{\vec{0}\} \text{ and }
  \trueclassifier(\vec{x}) = 1$.

\paragraph{Lower-bound constraint.}
We obtain $c_\ell$ by adding to the NN encoding of $g$ a new final layer with weight
and bias vectors
\begin{gather*}
  \vec{W} = 
  \begin{pmatrix}
    0\\
    1
  \end{pmatrix},
  \vec{B} =
  \begin{pmatrix}
    \ell\\
    0
  \end{pmatrix}
\end{gather*}
to obtain the NN-encoded function $\underline{g} \colon \Reals^n \to
\Reals^2$. Note that $\underline{g}(\vec{v}) = (\ell,g(\vec{v}))^\intercal$
for all $\vec{v} \in \Reals^n$. It follows that $c_\ell(\vec{v}) = 2$ if
$g(\vec{v}) > \ell$ and $c_\ell(\vec{v})$ is undefined if $g(\vec{v}) = \ell$.
In all other cases the classifier yields $1$.

\paragraph{Upper-bound constraint.}
To obtain $c_u$ we add to the NN encoding of $g$ two new layers. The
corresponding weight matrices and bias vectors are as follows.
\begin{align*}
  \vec{W}^{(1)} = (1),\ \vec{B}^{(1)} = (-u),\ 
  \vec{W}^{(2)} = 
  \begin{pmatrix}
    0\\
    -1%
	\end{pmatrix},\ 
	\vec{B}^{(2)} = 
  \begin{pmatrix}
    1\\
    1
  \end{pmatrix}
\end{align*}
Let us denote by $\overline{g}\colon \Reals^n \to \Reals^2$ the resulting
function. Observe that $\overline{g}(\vec{v}) = (1, \max(0, 1 -
\max(0,g(\vec{v})-u)))^\intercal$ for all $\vec{v} \in \Reals^n$. Hence, we have
that $c_u(\vec{v})$ is undefined if and only if $g(\vec{v}) \leq u$ and yields
$1$ otherwise.

\begin{lemma}\label{lem:intervals}
  Let $g \colon \Reals^n \to \Reals$ be an NN-encoded function and consider
  the constraint $\ell \leq g(\vec{x}) \leq u$. One can construct NN-encoded
  classifiers $c_\ell$ and $c_u$, of size linear with respect to $g$, for the
  labelled data set $(\{\vec{0}\},\{\vec{0} \mapsto 1\})$ such that the
  deterministic attack $\attack \colon \vec{0} \mapsto \vec{v}$
  \begin{itemize}
    \item induces a misclassification of $\vec{0}$ with respect to $c_\ell$ if
      and only if $\ell \leq g(\vec{v})$ and
    \item it induces a misclassification of $\vec{0}$ with respect to $c_u$ if
      and only if $g(\vec{v}) \leq u$.
  \end{itemize}
\end{lemma}
We now show how to modify the NN to obtain classifiers
$c_\ell,c_u$ such that $\vec{x}$ is misclassified by both $c_\ell$ and $c_u$
if and only if the constraint holds.

\subsubsection{Enforcing universal misclassification.}\label{sec:all-mis}
A randomized attack with misclassification value $1$ can be
assumed to be deterministic. 
Indeed, from Equation~\eqref{eqn:simple} it
follows that for any such randomized attack we must have $M_c(\attack) = |X|$
for all $\attack \in A$ and all $c \in C$. Hence, we can choose any such
$\attack \in A$ and consider the randomized attack $(\{\attack\}, \{\attack
\mapsto 1\})$ which also has misclassification value $1$.
\begin{lemma}\label{lem:all-mis}
  Consider the labelled data set $\datasetLabelled = (X,\trueclassifier)$ with the
  finite set of classifiers $\classifiers \subseteq \allClassifiers$. There
  exists an $\epsilon$-bounded randomized attack $(\attackSet,\probDistr)$ with
  $\Val(\attackSet,\probDistr)=1$ if and only if there exists a deterministic attack
  $\attack$ such that
  \begin{itemize}
    \item $\norm{\attack(\vec{x})}_1 \leq \epsilon$ for all $\vec{x} \in
      \dataset$ and
    \item for all $\vec{x} \in \dataset$ and all $c \in \classifiers$ we have that
      either $c(\vec{x}+\attack(\vec{x}))$ is undefined or it is not equal to $\trueclassifier(\vec{x})$.
  \end{itemize}
\end{lemma}

With Lemmas~\ref{lem:intervals} and~\ref{lem:all-mis} established, the proof of Theorem~\ref{theorem:np} is now complete.

\section{SMT and MILP Encodings}\label{sec:encodings}
In this section, we describe the main elements of our SMT and MILP encodings to compute (optimal) randomized attacks or prove the robustness of classifier ensembles.
We start with a base encoding and will afterwards explain how to explicitly encode the classifiers and the loss function.

\subsection{Base problem encoding}
First, we assume a labelled data set $\datasetLabelled=(\dataset,\trueclassifier)$, the attack bound $\epsilon$, and the robustness bound $\alpha$ are input to the problem.
In particular, the data set $\dataset=\{\datapoint^{1},\ldots,\datapoint^{|\dataset|}\}\subseteq\Reals^d$ has data points $\datapoint^{j}=(x_1^{j},\ldots,x_d^{j})\in\Reals^d$ for $1\leq j\leq |\dataset|$. 
Furthermore, we assume the number $|\attackSet|$ of attacks that shall be computed is fixed. 
Recall that, to show that the set of classifiers is robust, we can compute a sufficient bound on the number of attacks --- see Sec.~\ref{sec:bounded_number_of_attacks}.

For readability, we assume the classifiers $\classifiers$ and the loss function $\lossZeroOne$ are given as functions that can directly be used in the encodings, and
we will use the absolute value $|x|$ for $x\in\Reals$.
Afterwards, we discuss how to actually encode classifiers and the loss function.
We use the following variables:
\begin{itemize}
		\item For the attacks from $A$, we introduce $\attack_1,\ldots,\attack_{|\attackSet|}$ with $\attack_i\in\Reals^{|\dataset|\times d}$ for $1\leq i\leq |\attackSet|$. Specifically, $\attack_i$ shall be assigned all attack values for the $i$-th attack from $\attackSet$. That is, $\attack_i^j$ is the attack for the data point $\datapoint^j=(x_1^{j},\ldots,x_d^{j})\in\Reals^d$ with $\attack_i^j=(\attack_i^{j,1},\ldots,\attack_i^{j,d})$ for $1\leq j\leq |\dataset|$.		
	\item We introduce $\prob_1,\ldots,\prob_{|\attackSet|}$ to form a probability distribution over deterministic attacks; $\prob_i$ is assigned the probability to execute attack $\delta_{i}$.
\end{itemize}
The classifier ensemble $\classifiers$ is not $\epsilon,\alpha$-robust as in Definition~\ref{def:robustness} if and only if the following constraints are satisfiable.
\begin{align}
	\forall \classifier\in\classifiers.\quad & \sum_{j=1}^{|\dataset|} \sum_{i=1}^{|\attackSet|} \left(\prob_i\cdot\lossZeroOne(\classifier,\datapoint^{j},\attack_{i}^j)\right) \geq \alpha\cdot |\dataset|\label{eq:base_smt_misclassification_value}\\
	\forall i \in \{1,\dots,|\attackSet|\}, j \in \{1,\dots,|\dataset|\}. \quad & \sum_{k=1}^{d}|\attack^{j,k}_i|\leq \epsilon\label{eq:base_smt_epsilon}\\
	& \sum_{i=1}^{|\attackSet|}\prob_i=1\label{eq:base_smt_prob1}\\
	\forall i\in \{1,\dots,|\attackSet|\}.\quad & \prob_i\geq 0\label{eq:base_smt_prob_ge0}
\end{align}
Indeed, \eqref{eq:base_smt_misclassification_value} enforces the misclassification value to be at least $\alpha$;
\eqref{eq:base_smt_epsilon} ensures an $\epsilon$-bounded randomized
attack; finally, by~\eqref{eq:base_smt_prob1} and~\eqref{eq:base_smt_prob_ge0} the probability
variables induce a valid probability distribution.

\paragraph{Specific encodings.}
For the SMT encoding, we can make use of the $\max(\cdot)$ native to implement the absolute value. 
In particular for the MILP, however, we employ so-called ``big-M'' tricks to encode max functions and a (restricted) product operation (cf.~\cite{DBLP:conf/nips/BunelTTKM18}).
Specifically, the product is required to obtain the value resulting from the multiplication of the loss function and probability variables.

As an example of ``big-M'' trick, suppose we have variables $a \in \mathbb{Q} \cap [0,1], b \in \{0,1\}$, and a constant $M \in \mathbb{Q}$ such that $M > a + b$. We introduce a variable
$c \in \mathbb{Q} \cap [0,1]$ and add the following constraints which
clearly enforce that $c = ab$.
\begin{align*}
c &\geq a - M(1-b),\quad c \leq a + M(1-b),\quad c \leq 0 + Mb\ .
\end{align*}
Note that $M$ can be chosen to be the constant $2$ in this case.

\paragraph{Encoding the loss function.}
We encode the zero-one loss function from Def.~\ref{def:zero_one_loss} as an if-then-else expression making use of the ground truth classifier $\trueclassifier$.
We introduce one variable $\ell^c_{i,j}$
per classifier $\classifier\in\classifiers$
for all attacks
$\attack_i \in \attackSet$ and all datapoints
$\datapoint^j\in\dataset$. In the SMT encoding,
we can then define
\begin{align}
	 \ell^c_{i,j}=\mathit{ITE}((\classifier(\datapoint^j+\attack^j_i)=\trueclassifier(\datapoint^j)),0,1)
	\label{eq:smt_loss_function}
\end{align}
so that $\ell^c_{i,j} = \lossZeroOne(\classifier,\datapoint^{j},\attack_{i}^j)$. In our MILP encoding we have to simulate the \textit{ITE}
primitive using constraints similar to the ones mentioned above.

\subsection{Classifier encoding}

As mentioned in the preliminaries, neural networks implement
functions by way of layer composition. 
Intuitively, the input of a layer is by a previous layer.
When fed forward, input values are multiplied by a weight,
and a bias value will be added to it.
%
%
Matrix operations realized by a neural network can thus be encoded as linear functions. For max-pooling operations and the ReLU activation function, one can use the native $\max(\cdot)$ operation or implement a maximum using a ``big-M trick''. For this, a suitable constant $M$ has to be obtained beforehand (cf.~\cite{DBLP:conf/nips/BunelTTKM18}). We also use a (discrete) convolution operation, as a linear function.

\FloatBarrier
\section{Experiments}
In the previous section, we showed that our problem of neural network robustness verification is \NP-hard.
Meaningful comparison between approaches, therefore, needs to be experimental. 
To that end, we use classifiers trained on multiple image data sets and report on the comparison between the SMT and MILP encodings.
In what follows, we analyze the running time behavior of the different verifiers, the generated attacks and the misclassification value for the given data points, and whether a set of classifiers is robust against predefined thresholds.

\subsection{Experimental setup}
For each experiment, we define a set of classifiers $\classifiers$, our data points $\datasetLabelled$, the number of attacks $|\attackSet|$, and both the $\epsilon$- and $\alpha$-values. 
Then, we generate the attacks $\attackSet$ and the probability distribution $\probDistr$ using SMT and MILP solvers. 
If no randomized attack $(\attackSet,\probDistr)$ is found (UNSAT), we have shown that our set of classifiers is $\epsilon,\alpha$-robust with respect to the predefined thresholds.

\paragraph*{Toolchain.} Our NN robustness verifier\footnote{Available at \url{https://tinyurl.com/ensemble-robustness}.}, is implemented as part of a Python 3.x toolchain.
We use the SMT solver Z3~\cite{de2008z3} and the MILP solver Gurobi~\cite{gurobi} with their
standard settings.
To support arbitrary classifiers, we created a generic pipeline using the TensorFlow API, and support Max-Pooling layers, convolutional layers, and dense layers with ReLU activation functions~\cite{abadi16}.
We focus on binary classifiers by defining certain classification boundaries.
We train the classifiers using the Adam optimizer~\cite{kb14} as well as stochastic gradient descent.

\paragraph*{Data sets.} 
\textit{MNIST} consists of 70\,000 images of handwritten digits)~\cite{deng2012mnist} and is widely used for benchmarking in the field of machine learning~\cite{keysers2007comparison,cats17}.
We trained classifiers to have a test accuracy of at least 90\%.

\textit{German traffic sign} is a multi-class and single-image classification data set, containing more than 50\,000 images and more than 40 classes~\cite{stallkamp2011german}.
Traffic sign recognition and potential attacks are of utmost importance for self-driving cars~\cite{yan2016can}.
We extracted the images for ``Give way'' and ``priority'' traffic signs from the data set and trained classifiers on this subset to have an accuracy of at least 80\%.

\begin{table}[t]
\centering\scalebox{0.8}{
\begin{tabular}{|l|l|l|l|l|c|l|l|l|l|l|l|}
\hline
\multicolumn{8}{|l|}{{\color[HTML]{000000} \textbf{Benchmark Information}}}                                                                                                                                                                                                                                                 & \multicolumn{2}{l|}{{\color[HTML]{000000} \textbf{SMT}}}                        & \multicolumn{2}{l|}{{\color[HTML]{000000} \textbf{MILP}}}                       \\ \hline
{\color[HTML]{000000} \textbf{ID}} & {\color[HTML]{000000} \textbf{Name}}        & {\color[HTML]{000000} \textbf{$|\boldsymbol{\classifiers}|$}} & {\color[HTML]{000000} \textbf{$|\boldsymbol{\attackSet}|$}} & {\color[HTML]{000000} \textbf{$|\boldsymbol{\datasetLabelled}|$}} & {\color[HTML]{000000} \textbf{dim}} & {\color[HTML]{000000} \textbf{$\boldsymbol{\alpha}$}} & {\color[HTML]{000000} \textbf{$\boldsymbol{\epsilon}$}} & {\color[HTML]{000000} \textbf{Time}} & {\color[HTML]{000000} \textbf{$\boldsymbol{\Val(\attackSet,\probDistr)}$}} & {\color[HTML]{000000} \textbf{Time}} & {\color[HTML]{000000} \textbf{$\boldsymbol{\Val(\attackSet,\probDistr)}$}}  \\ \hline
2           & mnist\_0\_1         & 3            & 2            & 4            & 7x7           & 0.2            & 100              & -TO-          & --                & 12.43          & 0.25              \\ \hline
4           & mnist\_0\_1\_2convs & 3            & 2            & 4            & 8x8           & 0.4            & 100              & -TO-          & --                & 53.92          & 0.4               \\ \hline
7           & mnist\_0\_1         & 3            & 2            & 4            & 8x8           & 0.4            & 1000             & -TO-          & --                & 0.34           & 0.4               \\ \hline
8           & mnist\_0\_1         & 3            & 2            & 4            & 8x8           & 0.9            & 1000             & -TO-          & --                & 50.09          & 0.9               \\ \hline
9           & mnist\_0\_1         & 3            & 3            & 4            & 8x8           & 0.9            & 60               & -TO-          & --                & 34.02          & 1                 \\ \hline
13          & mnist\_4\_5         & 3            & 4            & 4            & 10x10         & 0.9            & 50               & -TO-          & --                & 144.32         & 1                 \\ \hline
14          & mnist\_7\_8         & 3            & 4            & 4            & 6x6           & 0.1            & 60               & -TO-          & --                & 18.94          & 0.25              \\ \hline
16          & mnist\_4\_5         & 3            & 2            & 4            & 10x10         & 0.1            & 1000             & 155.73        & 0.38              & 101.16         & 0.1               \\ \hline
17          & mnist\_4\_5         & 3            & 3            & 4            & 10x10         & 0.1            & 80               & 403.25        & 0.25              & 101.47         & 0.25              \\ \hline
18          & mnist\_4\_5         & 3            & 2            & 4            & 10x10         & 0.15           & 80               & 216.65        & 0.38              & 44.26          & 0.15              \\ \hline
19          & mnist\_4\_5         & 3            & 2            & 4            & 10x10         & 0.2            & 100              & 156.63        & 0.38              & 54.36          & 0.25              \\ \hline
22          & mnist\_7\_8         & 3            & 2            & 4            & 6x6           & 0.9            & 0.1              & -TO-          & --                & 4              & robust            \\ \hline
26          & traffic\_signs      & 3            & 2            & 4            & 10x10         & 1              & 0.01             & -TO-          & --                & 17             & robust            \\ \hline
27          & traffic\_signs      & 3            & 2            & 4            & 10x10         & 1              & 0.1              & -TO-          & --                & -TO-           & --                \\ \hline
\end{tabular}
}
\caption{SMT versus MILP}
\label{tab:smt_milp}
\end{table}

\begin{table}[t]
\centering\scalebox{0.8}{
\begin{tabular}{|l|l|l|l|l|c|l|l|l|l|l|l|}
\hline
\multicolumn{8}{|l|}{{\color[HTML]{000000} \textbf{Benchmark Information}}}                                                                                                                                                                                                                                                 & \multicolumn{2}{l|}{{\color[HTML]{000000} \textbf{MILP}}}                       & \multicolumn{2}{l|}{{\color[HTML]{000000} \textbf{MaxMILP}}}                    \\ \hline
{\color[HTML]{000000} \textbf{ID}} & {\color[HTML]{000000} \textbf{Name}}        & {\color[HTML]{000000} \textbf{$|\boldsymbol{\classifiers|}$}} & {\color[HTML]{000000} \textbf{$|\boldsymbol{\attackSet|}$}} & {\color[HTML]{000000} \textbf{$|\boldsymbol{\datasetLabelled}|$}} & {\color[HTML]{000000} \textbf{dim}} & {\color[HTML]{000000} \textbf{$\boldsymbol{\alpha}$}} & {\color[HTML]{000000} \textbf{$\boldsymbol{\epsilon}$}} & {\color[HTML]{000000} \textbf{Time}} & {\color[HTML]{000000} \textbf{$\boldsymbol{\Val(\attackSet,\probDistr)}$}} & {\color[HTML]{000000} \textbf{Time}} & {\color[HTML]{000000} \textbf{$\boldsymbol{\Val(\attackSet,\probDistr)}$}} \\ \hline
1           & mnist\_0\_1         & 3            & 2            & 4            & 7x7           & 0.1            & 1000             & 57.79          & 0.25              & 46.23           & 1*                  \\ \hline
3           & mnist\_0\_1\_2convs & 3            & 2            & 4            & 8x8           & 0.2            & 1000             & 738.76         & 0.5               & 93.54           & 1*                  \\ \hline
7           & mnist\_0\_1         & 3            & 2            & 4            & 8x8           & 0.4            & 1000             & 0.34           & 0.4               & 0.34            & 1*                  \\ \hline
10          & mnist\_0\_1         & 3            & 4            & 4            & 8x8           & 0.9            & 60               & 51.39          & 1                 & 51.39           & 1*                  \\ \hline
14          & mnist\_7\_8         & 3            & 4            & 4            & 6x6           & 0.1            & 60               & 18.94          & 0.25              & 21.20           & 1                   \\ \hline
17          & mnist\_4\_5         & 3            & 3            & 4            & 10x10         & 0.1            & 80               & 101.47         & 0.25              & 88.39           & 1                   \\ \hline
20          & mnist\_3\_6         & 2            & 9            & 2            & 8x8           & 1              & 0.005            & 7              & robust            & 7               & robust              \\ \hline
21          & mnist\_7\_8         & 3            & 2            & 4            & 6x6           & 1              & 0.1              & 4              & robust            & 4               & robust              \\ \hline
24          & mnist\_0\_2         & 3            & 27           & 4            & 9x9           & 1              & 0.01             & 108            & robust            & 108             & robust              \\ \hline
25          & mnist\_0\_2         & 3            & 30           & 4            & 9x9           & 1              & 0.01             & 120            & robust            & 120             & robust              \\ \hline
28          & traffic\_signs      & 3            & 3            & 4            & 10x10         & 1              & 0.01             & 45             & robust            & 45              & robust              \\ \hline
29          & traffic\_signs      & 3            & 3            & 4            & 10x10         & 1              & 0.01             & --TO--         & -                 & --TO--          & --                  \\ \hline
\end{tabular}
}
\caption{MILP  versus  MaxMILP}
\label{tab:milp_mmilp}
\end{table}

\paragraph{Optimal attacks.}
As MILP inherently solves optimization problems, we augment the base encoding from Equations~\eqref{eq:base_smt_misclassification_value}--\eqref{eq:base_smt_prob_ge0} with the following objective function:
\[
    \max\quad\sum_{k=1}^{|\classifiers|} \sum_{j=1}^{|\datasetLabelled|} \sum_{i=1}^{|\attackSet|} \left(\prob_i\cdot\lossZeroOne(\classifier_k,\datapoint^{j},\attack_{i}^j)\right).
\]
An optimal solution with respect to the objective \textbf{may} 
yield a randomized attack inducing the maximal misclassification value among all $\epsilon$-bounded attacks.\footnote{Note that we sum over classifiers instead of minimizing, as required in $\Val(\attackSet,\probDistr)$.}

\paragraph*{Alternative attacks.}
Our method generates attacks taking the whole ensemble of classifiers into account, which is computationally harder than just considering single classifiers~\cite{DBLP:conf/cav/KatzBDJK17,DBLP:conf/atva/Ehlers17} due to an increased number of constraints and variables in the underlying encodings.
To showcase the need for our approach, we implemented two other ways to generate attacks that are based on individual classifiers and subsequently lifted to the whole ensemble.
Recall that we assume the attacker does not know which classifier from the ensemble will be chosen.

First, for the classifier ensemble $\classifiers$ we compute --- using a simplified version of our MILP encoding --- an optimal attack $\attack_\classifier$ for each classifier $\classifier\in\classifiers$. 
Each such $\attack_\classifier$ maximizes the misclassification value, that is, the loss, for the classifier $\classifier$. 
The attack set $\attackSet_\classifiers=\{\attack_\classifier\mid\classifier\in\classifiers\}$
together with a uniform distribution $\Distr_\attackSet$ over $\attackSet_\classifiers$
form the so-called \emph{uniform attacker} $(\attackSet_\classifiers,\Distr_\attackSet)$.

Second, to compare with deterministic attacks, we calculate for each attack from $\attackSet_\classifiers$ the misclassification value over all classifiers. 
The \emph{best deterministic attack}
is any attack from 
$\attackSet_\classifiers$ inducing a maximal misclassification value.


\subsection{Evaluation}
We report on our experimental results using the aforementioned data sets and attacks. 
For all experiments we used a timeout of 7200 seconds (TO).
Each benchmark has an ID, a name, the number $|\classifiers|$ of classifiers, the number $|\attackSet|$ of attacks, the size $|\datasetLabelled|$ of the data set, 
the dimension of the image (dim), the robustness bound $\alpha$, and the attack bound $\epsilon$. 
The names of the MNIST benchmarks are of the form ``mnist\_$x$\_$y$'', where $x$ and $y$ are the labels; the additional suffix ``\_$n$convs'' indicates that the classifier has $n$ convolutional layers.
We provide an excerpt of our experiments, full tables are available in the appendix.

%
\paragraph*{SMT versus MILP.}
In Table~\ref{tab:smt_milp}, we report on the comparison between SMT and MILP.
Note that for these benchmarks, the MILP solver just checks the feasibility of the constraints without an objective function.
We list for both solvers the time in seconds and the misclassification value 
$\Val(\attackSet,\probDistr)$, rounded to 2 decimal places, for the generated randomized attack $(\attackSet,\probDistr)$, if it can be computed before the timeout.
If there is no solution, the classifier ensemble $\classifiers$ is $\epsilon,\alpha$-robust, and instead of a misclassification value we list ``robust''.

We observe that SMT is only able to find solutions within the timeout for small $\alpha$ and large $\epsilon$ values.
Moreover, if the ensemble is robust, that is, the problem is not satisfiable, the solver does not terminate for any benchmark. {Nevertheless, for some
benchmarks (see Table~\ref{tab:smt_milp}, entries 16--19) the SMT solver yields a higher misclassification value than 
the MILP solver --- that is,
it finds a better attack.}
The MILP solver, on the other hand, solves most of our benchmarks mostly within less than a minute, including the robust ones.
Despite the reasonably low timeout of 7200 seconds, we are thus able to verify the robustness of NNs with around 6 layers.
Running times visibly differ for other factors such as layer types.

\paragraph*{MILP versus MaxMILP.} 
Table~\ref{tab:milp_mmilp} compares some of the MILP results to those where we optimize the mentioned objective function, denoted by MaxMILP.
The MILP solver Gurobi offers the possibility of so-called \emph{callbacks}, that is, while an intermediate solution is not proven to be optimal, it may already be feasible. 
In case optimality cannot be shown within the timeout, we list the current (feasible) solution, and mark optimal solutions with $*$.
The misclassification value for the MaxMILP solver is always 1.
For robust ensembles, it is interesting to see that the MaxMILP encoding sometimes needs less time.


\paragraph*{MaxMILP versus alternative attacks.} 

\begin{table}[t]
\centering
\scalebox{0.8}{
\begin{tabular}{|l|l|l|l|l|c|l|l|l|l|l|}
\hline
\multicolumn{8}{|l|}{{\color[HTML]{000000} \textbf{Benchmark Information}}}                                                                                                                                                                                                                                               & {\color[HTML]{000000} \textbf{UA}}       & {\color[HTML]{000000} \textbf{BDA}}      & {\color[HTML]{000000} \textbf{MaxMILP}}  \\ \hline
{\color[HTML]{000000} \textbf{ID}} & {\color[HTML]{000000} \textbf{Name}}       & {\color[HTML]{000000} \textbf{$|\boldsymbol{\classifiers|}$}} & {\color[HTML]{000000} \textbf{$|\boldsymbol{\attackSet|}$}} & {\color[HTML]{000000} \textbf{$|\boldsymbol{\datasetLabelled|}$}} & {\color[HTML]{000000} \textbf{dim}} & {\color[HTML]{000000} \textbf{Epsilon}} & {\color[HTML]{000000} \textbf{Alpha}} & {\color[HTML]{000000} \textbf{$\boldsymbol{\Val(\attackSet,\probDistr)}$}} & {\color[HTML]{000000} \textbf{$\boldsymbol{\Val(\attackSet,\probDistr)}$}} & {\color[HTML]{000000} \textbf{$\boldsymbol{\Val(\attackSet,\probDistr)}$}} \\ \hline
3           & mnist\_0\_1\_2convs & 3            & 2            & 4            & 8x8           & 0.2            & 1000             & 0.33              & 0.25               & 1*                \\ \hline
7           & mnist\_0\_1         & 3            & 2            & 4            & 8x8           & 0.4            & 1000             & 0.33              & 0.5               & 1*                \\ \hline
8           & mnist\_0\_1         & 3            & 2            & 4            & 8x8           & 0.9            & 1000             & 0.33              & 0.5               & 1*                \\ \hline
9           & mnist\_0\_1         & 3            & 3            & 4            & 8x8           & 0.9            & 60               & 0.33              & 0.5               & 1*                \\ \hline
10          & mnist\_0\_1         & 3            & 4            & 4            & 8x8           & 0.9            & 60               & 0.33              & 0.5               & 1*                \\ \hline
11          & mnist\_4\_5         & 3            & 2            & 4            & 10x10         & 0.2            & 100              & 0.33              & 0.25              & 1*                \\ \hline
14          & mnist\_7\_8         & 3            & 4            & 4            & 6x6           & 0.1            & 60               & 0.33              & 0.5               & 1                 \\ \hline
15          & mnist\_7\_8         & 3            & 10           & 4            & 6x6           & 0.1            & 60               & 0.33              & 0.5               & 1                 \\ \hline
16          & mnist\_4\_5         & 3            & 2            & 4            & 10x10         & 0.1            & 1000             & 0.33              & 0.75              & 1*                \\ \hline
17          & mnist\_4\_5         & 3            & 3            & 4            & 10x10         & 0.1            & 80               & 0.33              & 0.5               & 1                 \\ \hline
18          & mnist\_4\_5         & 3            & 2            & 4            & 10x10         & 0.15           & 80               & 0.33              & 0.5               & 1*                \\ \hline
\end{tabular}
}
\caption{Attacker Comparison}
\label{tab:attacker_comparison}
\end{table}
In Table~\ref{tab:attacker_comparison}, we compare the MaxMILP method to the \emph{uniform attacker} (UA) and the \emph{best deterministic attacker} (BDA). 
What we can see is that the best deterministic attacker usually achieves higher misclassification values than the uniform attacker, but none of them are able to reach the actual optimum of 1.

\paragraph{Discussion of the results.} Within the timeout, our method is able to generate optimal results for medium-sized neural networks. 
The running time is mainly influenced by the number and type of the used layers, in particular, it is governed by  convolutional and max-pooling layers: these involve more matrix operations than dense layers. As expected,
larger values of the robustness bound $\alpha$ and smaller
values of the attack bound $\epsilon$ typically increase the running times.

\FloatBarrier
\section{Conclusion and Future Work}
We presented a new method to formally verify the robustness or, vice versa, compute optimal attacks for an ensemble of classifiers. 
Despite the theoretical hardness, we were able to, in particular by using MILP-solving, provide results for meaningful benchmarks.
In future work, we will render our method more scalable towards a standalone verification tool for neural network ensembles.
Moreover, we will explore settings where we do not have white-box access to the classifiers and employ state-of-the-art classifier stealing methods.

\bibliographystyle{splncs04}
\bibliography{literature}

\begin{thebibliography}{10}
\providecommand{\url}[1]{\texttt{#1}}
\providecommand{\urlprefix}{URL }
\providecommand{\doi}[1]{https://doi.org/#1}

\bibitem{abadi16}
Abadi, M.: Tensorflow: learning functions at scale. In: Garrigue, J., Keller,
  G., Sumii, E. (eds.) ICFP. p.~1. {ACM} (2016)

\bibitem{DBLP:conf/iclr/AbbasiG17}
Abbasi, M., Gagn{\'{e}}, C.: Robustness to adversarial examples through an
  ensemble of specialists. In: {ICLR} (Workshop). OpenReview.net (2017)

\bibitem{abbasi2020toward}
Abbasi, M., Rajabi, A., Gagn{\'{e}}, C., Bobba, R.B.: Toward adversarial
  robustness by diversity in an ensemble of specialized deep neural networks.
  In: Canadian Conference on {AI}. LNCS, vol. 12109, pp. 1--14. Springer (2020)

\bibitem{DBLP:conf/aaai/AkintundeKLP19}
Akintunde, M.E., Kevorchian, A., Lomuscio, A., Pirovano, E.: Verification of
  {RNN}-based neural agent-environment systems. In: {AAAI}. pp. 6006--6013.
  {AAAI} Press (2019)

\bibitem{amodei2016concrete}
Amodei, D., Olah, C., Steinhardt, J., Christiano, P., Schulman, J., Man{\'e},
  D.: Concrete problems in ai safety. CoRR  \textbf{abs/1606.06565} (2016)

\bibitem{ag11}
Apt, K.R., Gr{\"a}del, E.: Lectures in game theory for computer scientists.
  Cambridge University Press (2011)

\bibitem{DBLP:conf/nips/BunelTTKM18}
Bunel, R., Turkaslan, I., Torr, P.H.S., Kohli, P., Mudigonda, P.K.: A unified
  view of piecewise linear neural network verification. In: NeurIPS. pp.
  4795--4804 (2018)

\bibitem{cats17}
Cohen, G., Afshar, S., Tapson, J., van Schaik, A.: {EMNIST:} extending {MNIST}
  to handwritten letters. In: IJCNN. pp. 2921--2926. {IEEE} (2017)

\bibitem{de2008z3}
De~Moura, L., Bj{\o}rner, N.: Z3: An efficient smt solver. In: TACAS. pp.
  337--340. Springer (2008)

\bibitem{deng2012mnist}
Deng, L.: The {MNIST} database of handwritten digit images for machine learning
  research [best of the web]. {IEEE} Signal Process. Mag.  \textbf{29}(6),
  141--142 (2012)

\bibitem{sanjit-robust}
Dreossi, T., Ghosh, S., Sangiovanni{-}Vincentelli, A.L., Seshia, S.A.: A
  formalization of robustness for deep neural networks. CoRR
  \textbf{abs/1903.10033} (2019)

\bibitem{DBLP:conf/atva/Ehlers17}
Ehlers, R.: Formal verification of piece-wise linear feed-forward neural
  networks. In: ATVA. LNCS, vol. 10482, pp. 269--286. Springer (2017)

\bibitem{freedman2016safety}
Freedman, R.G., Zilberstein, S.: Safety in {AI-HRI}: Challenges complementing
  user experience quality. In: AAAI Fall Symposium Series (2016)

\bibitem{gurobi}
{Gurobi Optimization, Inc.}: Gurobi optimizer reference manual.
  \url{http://www.gurobi.com} (2013)

\bibitem{DBLP:conf/cav/KatzBDJK17}
Katz, G., Barrett, C.W., Dill, D.L., Julian, K., Kochenderfer, M.J.: Reluplex:
  An efficient {SMT} solver for verifying deep neural networks. In: CAV. LNCS,
  vol. 10426, pp. 97--117. Springer (2017)

\bibitem{keysers2007comparison}
Keysers, D.: Comparison and combination of state-of-the-art techniques for
  handwritten character recognition: topping the mnist benchmark. arXiv
  preprint arXiv:0710.2231  (2007)

\bibitem{kb14}
Kingma, D.P., Ba, J.: Adam: {A} method for stochastic optimization. In: Bengio,
  Y., LeCun, Y. (eds.) ICLR (2015), \url{http://arxiv.org/abs/1412.6980}

\bibitem{DBLP:conf/concur/Kwiatkowska19}
Kwiatkowska, M.Z.: Safety verification for deep neural networks with provable
  guarantees (invited paper). In: CONCUR. LIPIcs, vol.~140, pp. 1:1--1:5.
  Schloss Dagstuhl - Leibniz-Zentrum f{\"{u}}r Informatik (2019)

\bibitem{nyholm2018the}
{Nyholm}, S.: The ethics of crashes with self-driving cars: A roadmap, ii.
  Philosophy Compass  \textbf{13}(7) (2018)

\bibitem{DBLP:journals/corr/abs-1906-02816}
Perdomo, J.C., Singer, Y.: Robust attacks against multiple classifiers. CoRR
  \textbf{abs/1906.02816} (2019)

\bibitem{pinot2020randomization}
Pinot, R., Ettedgui, R., Rizk, G., Chevaleyre, Y., Atif, J.: Randomization
  matters. how to defend against strong adversarial attacks. CoRR
  \textbf{abs/2002.11565} (2020)

\bibitem{ranzatorobustness}
Ranzato, F., Zanella, M.: Robustness verification of decision tree ensembles.
  In: OVERLAY@AI*IA. vol.~2509, pp. 59--64. CEUR-WS.org (2019)

\bibitem{future_AI}
Science, N., (NSTC), T.C.: Preparing for the Future of Artificial Intelligence
  (2016)

\bibitem{popl-vechev}
Singh, G., Gehr, T., P{\"{u}}schel, M., Vechev, M.T.: An abstract domain for
  certifying neural networks. Proc. {ACM} Program. Lang.  \textbf{3}({POPL}),
  41:1--41:30 (2019). \doi{10.1145/3290354},
  \url{https://doi.org/10.1145/3290354}

\bibitem{stallkamp2011german}
Stallkamp, J., Schlipsing, M., Salmen, J., Igel, C.: The german traffic sign
  recognition benchmark: a multi-class classification competition. In: IJCNN.
  pp. 1453--1460. IEEE (2011)

\bibitem{stoica2017berkeley}
Stoica, I., Song, D., Popa, R.A., Patterson, D., Mahoney, M.W., Katz, R.,
  Joseph, A.D., Jordan, M., Hellerstein, J.M., Gonzalez, J.E., et~al.: A
  {B}erkeley view of systems challenges for {AI}. CoRR  \textbf{abs/1712.05855}
  (2017)

\bibitem{yan2016can}
Yan, C., Xu, W., Liu, J.: Can you trust autonomous vehicles: Contactless
  attacks against sensors of self-driving vehicle. DEF CON  \textbf{24}(8),
  ~109 (2016)

\end{thebibliography}

\clearpage

\appendix
\section{Full Experimental Results}

\begin{table}[]
\begin{tabular}{|l|l|l|l|l|c|l|l|l|l|l|l|}
\hline
\multicolumn{8}{|l|}{{\color[HTML]{000000} \textbf{Benchmark Information}}}                                                                                                                                                                                                                                                 & \multicolumn{2}{l|}{{\color[HTML]{000000} \textbf{SMT}}}                        & \multicolumn{2}{l|}{{\color[HTML]{000000} \textbf{MILP}}}                       \\ \hline
{\color[HTML]{000000} \textbf{ID}} & {\color[HTML]{000000} \textbf{Name}}        & {\color[HTML]{000000} \textbf{$|\boldsymbol{\classifiers}|$}} & {\color[HTML]{000000} \textbf{$|\boldsymbol{\attackSet}|$}} & {\color[HTML]{000000} \textbf{$|\boldsymbol{\datasetLabelled}|$}} & {\color[HTML]{000000} \textbf{dim}} & {\color[HTML]{000000} \textbf{$\boldsymbol{\alpha}$}} & {\color[HTML]{000000} \textbf{$\boldsymbol{\epsilon}$}} & {\color[HTML]{000000} \textbf{Time}} & {\color[HTML]{000000} \textbf{$\boldsymbol{\Val(\attackSet,\probDistr)}$}} & {\color[HTML]{000000} \textbf{Time}} & {\color[HTML]{000000} \textbf{$\boldsymbol{\Val(\attackSet,\probDistr)}$}}  \\ \hline
1           & mnist\_0\_1         & 3            & 2            & 4            & 7x7           & 0.1            & 1000             & -TO-          & --                & 57.79          & 0.25              \\ \hline
2           & mnist\_0\_1         & 3            & 2            & 4            & 7x7           & 0.2            & 100              & -TO-          & --                & 12.43          & 0.25              \\ \hline
3           & mnist\_0\_1\_2convs & 3            & 2            & 4            & 8x8           & 0.2            & 1000             & -TO-          & --                & 738.76         & 0.5               \\ \hline
4           & mnist\_0\_1\_2convs & 3            & 2            & 4            & 8x8           & 0.4            & 100              & -TO-          & --                & 53.92          & 0.4               \\ \hline
5           & mnist\_0\_1\_2convs & 3            & 2            & 4            & 8x8           & 0.9            & 100              & -TO-          & --                & 44.24          & 1                 \\ \hline
6           & mnist\_0\_1\_2convs & 3            & 3            & 4            & 8x8           & 0.9            & 100              & -TO-          & --                & 96.78          & 1                 \\ \hline
7           & mnist\_0\_1         & 3            & 2            & 4            & 8x8           & 0.4            & 1000             & -TO-          & --                & 0.34           & 0.4               \\ \hline
8           & mnist\_0\_1         & 3            & 2            & 4            & 8x8           & 0.9            & 1000             & -TO-          & --                & 50.09          & 0.9               \\ \hline
9           & mnist\_0\_1         & 3            & 3            & 4            & 8x8           & 0.9            & 60               & -TO-          & --                & 34.02          & 1                 \\ \hline
10          & mnist\_0\_1         & 3            & 4            & 4            & 8x8           & 0.9            & 60               & -TO-          & --                & 51.39          & 1                 \\ \hline
11          & mnist\_4\_5         & 3            & 2            & 4            & 10x10         & 0.2            & 100              & -TO-          & --                & 57.92          & 0.25              \\ \hline
12          & mnist\_4\_5         & 3            & 4            & 4            & 10x10         & 0.3            & 50               & -TO-          & --                & 143.86         & 0.5               \\ \hline
13          & mnist\_4\_5         & 3            & 4            & 4            & 10x10         & 0.9            & 50               & -TO-          & --                & 144.32         & 1                 \\ \hline
14          & mnist\_7\_8         & 3            & 4            & 4            & 6x6           & 0.1            & 60               & -TO-          & --                & 18.94          & 0.25              \\ \hline
15          & mnist\_7\_8         & 3            & 10           & 4            & 6x6           & 0.1            & 60               & -TO-          & --                & 76.25          & 0.5               \\ \hline
16          & mnist\_4\_5         & 3            & 2            & 4            & 10x10         & 0.1            & 1000             & 155.73        & 0.38              & 101.16         & 0.1               \\ \hline
17          & mnist\_4\_5         & 3            & 3            & 4            & 10x10         & 0.1            & 80               & 403.25        & 0.25              & 101.47         & 0.25              \\ \hline
18          & mnist\_4\_5         & 3            & 2            & 4            & 10x10         & 0.15           & 80               & 216.65        & 0.38              & 44.26          & 0.15              \\ \hline
19          & mnist\_4\_5         & 3            & 2            & 4            & 10x10         & 0.2            & 100              & 156.63        & 0.38              & 54.36          & 0.25              \\ \hline
20          & mnist\_3\_6         & 2            & 9            & 2            & 8x8           & 1              & 0.005            & -TO-          & --                & 7              & robust            \\ \hline
21          & mnist\_7\_8         & 3            & 2            & 4            & 6x6           & 1              & 0.1              & -TO-          & --                & 4              & robust            \\ \hline
22          & mnist\_7\_8         & 3            & 2            & 4            & 6x6           & 0.9            & 0.1              & -TO-          & --                & 4              & robust            \\ \hline
23          & mnist\_7\_8         & 3            & 7            & 4            & 6x6           & 0.9            & 0.3              & -TO-          & --                & 71             & robust            \\ \hline
24          & mnist\_0\_2         & 3            & 27           & 4            & 9x9           & 1              & 0.01             & -TO-          & --                & 108            & robust            \\ \hline
25          & mnist\_0\_2         & 3            & 30           & 4            & 9x9           & 1              & 0.01             & -TO-          & --                & 120            & robust            \\ \hline
26          & traffic\_signs      & 3            & 2            & 4            & 10x10         & 1              & 0.01             & -TO-          & --                & 17             & robust            \\ \hline
27          & traffic\_signs      & 3            & 2            & 4            & 10x10         & 1              & 0.1              & -TO-          & --                & -TO-           & --                \\ \hline
28          & traffic\_signs      & 3            & 3            & 4            & 10x10         & 1              & 0.01             & -TO-          & --                & 45             & robust            \\ \hline
29          & traffic\_signs      & 3            & 3            & 4            & 10x10         & 1              & 0.01             & -TO-          & --                & -TO-           & --                \\ \hline
\end{tabular}
\caption{SMT versus MILP}
\label{tab:smt_milp_app}
\end{table}
\begin{table}[]
\begin{tabular}{|l|l|l|l|l|c|l|l|l|l|l|l|}
\hline
\multicolumn{8}{|l|}{{\color[HTML]{000000} \textbf{Benchmark Information}}}                                                                                                                                                                                                                                                 & \multicolumn{2}{l|}{{\color[HTML]{000000} \textbf{MILP}}}                       & \multicolumn{2}{l|}{{\color[HTML]{000000} \textbf{MaxMILP}}}                    \\ \hline
{\color[HTML]{000000} \textbf{ID}} & {\color[HTML]{000000} \textbf{Name}}        & {\color[HTML]{000000} \textbf{$|\boldsymbol{\classifiers|}$}} & {\color[HTML]{000000} \textbf{$|\boldsymbol{\attackSet|}$}} & {\color[HTML]{000000} \textbf{$|\boldsymbol{\datasetLabelled}|$}} & {\color[HTML]{000000} \textbf{dim}} & {\color[HTML]{000000} \textbf{$\boldsymbol{\alpha}$}} & {\color[HTML]{000000} \textbf{$\boldsymbol{\epsilon}$}} & {\color[HTML]{000000} \textbf{Time}} & {\color[HTML]{000000} \textbf{$\boldsymbol{\Val(\attackSet,\probDistr)}$}} & {\color[HTML]{000000} \textbf{Time}} & {\color[HTML]{000000} \textbf{$\boldsymbol{\Val(\attackSet,\probDistr)}$}} \\ \hline
1           & mnist\_0\_1         & 3            & 2            & 4            & 7x7           & 0.1            & 1000             & 57.79          & 0.25              & 46.23           & 1*                  \\ \hline
2           & mnist\_0\_1         & 3            & 2            & 4            & 7x7           & 0.2            & 100              & 12.43          & 0.25              & 12.43           & 1*                  \\ \hline
3           & mnist\_0\_1\_2convs & 3            & 2            & 4            & 8x8           & 0.2            & 1000             & 738.76         & 0.5               & 93.54           & 1*                  \\ \hline
4           & mnist\_0\_1\_2convs & 3            & 2            & 4            & 8x8           & 0.4            & 100              & 53.92          & 0.4               & 53.92           & 1*                  \\ \hline
5           & mnist\_0\_1\_2convs & 3            & 2            & 4            & 8x8           & 0.9            & 100              & 44.24          & 1                 & 44.24           & 1*                  \\ \hline
6           & mnist\_0\_1\_2convs & 3            & 3            & 4            & 8x8           & 0.9            & 100              & 96.78          & 1                 & 96.78           & 1*                  \\ \hline
7           & mnist\_0\_1         & 3            & 2            & 4            & 8x8           & 0.4            & 1000             & 0.34           & 0.4               & 0.34            & 1*                  \\ \hline
8           & mnist\_0\_1         & 3            & 2            & 4            & 8x8           & 0.9            & 1000             & 50.09          & 0.9               & 50.09           & 1*                  \\ \hline
9           & mnist\_0\_1         & 3            & 3            & 4            & 8x8           & 0.9            & 60               & 34.02          & 1                 & 34.02           & 1*                  \\ \hline
10          & mnist\_0\_1         & 3            & 4            & 4            & 8x8           & 0.9            & 60               & 51.39          & 1                 & 51.39           & 1*                  \\ \hline
11          & mnist\_4\_5         & 3            & 2            & 4            & 10x10         & 0.2            & 100              & 57.92          & 0.25              & 52.95           & 1*                  \\ \hline
12          & mnist\_4\_5         & 3            & 4            & 4            & 10x10         & 0.3            & 50               & 143.86         & 0.5               & 158.20          & 1                   \\ \hline
13          & mnist\_4\_5         & 3            & 4            & 4            & 10x10         & 0.9            & 50               & 144.32         & 1                 & 144.32          & 1                   \\ \hline
14          & mnist\_7\_8         & 3            & 4            & 4            & 6x6           & 0.1            & 60               & 18.94          & 0.25              & 21.20           & 1                   \\ \hline
15          & mnist\_7\_8         & 3            & 10           & 4            & 6x6           & 0.1            & 60               & 76.25          & 0.5               & 76.25           & 1                   \\ \hline
16          & mnist\_4\_5         & 3            & 2            & 4            & 10x10         & 0.1            & 1000             & 101.16         & 0.1               & 106.95          & 1*                  \\ \hline
17          & mnist\_4\_5         & 3            & 3            & 4            & 10x10         & 0.1            & 80               & 101.47         & 0.25              & 88.39           & 1                   \\ \hline
18          & mnist\_4\_5         & 3            & 2            & 4            & 10x10         & 0.15           & 80               & 44.26          & 0.15              & 52.00           & 1*                  \\ \hline
19          & mnist\_4\_5         & 3            & 2            & 4            & 10x10         & 0.2            & 100              & 54.36          & 0.25              & 44.00           & 1*                  \\ \hline
20          & mnist\_3\_6         & 2            & 9            & 2            & 8x8           & 1              & 0.005            & 7              & robust            & 7               & robust              \\ \hline
21          & mnist\_7\_8         & 3            & 2            & 4            & 6x6           & 1              & 0.1              & 4              & robust            & 4               & robust              \\ \hline
22          & mnist\_7\_8         & 3            & 2            & 4            & 6x6           & 0.9            & 0.1              & 4              & robust            & 4               & robust              \\ \hline
23          & mnist\_7\_8         & 3            & 7            & 4            & 6x6           & 0.9            & 0.3              & 71             & robust            & 71              & robust              \\ \hline
24          & mnist\_0\_2         & 3            & 27           & 4            & 9x9           & 1              & 0.01             & 108            & robust            & 108             & robust              \\ \hline
25          & mnist\_0\_2         & 3            & 30           & 4            & 9x9           & 1              & 0.01             & 120            & robust            & 120             & robust              \\ \hline
26          & traffic\_signs      & 3            & 2            & 4            & 10x10         & 1              & 0.01             & 17             & robust            & 17              & robust              \\ \hline
27          & traffic\_signs      & 3            & 2            & 4            & 10x10         & 1              & 0.1              & --TO--         & -                 & --TO--          & --                  \\ \hline
28          & traffic\_signs      & 3            & 3            & 4            & 10x10         & 1              & 0.01             & 45             & robust            & 45              & robust              \\ \hline
29          & traffic\_signs      & 3            & 3            & 4            & 10x10         & 1              & 0.01             & --TO--         & -                 & --TO--          & --                  \\ \hline
\end{tabular}
\caption{MILP  versus  MaxMILP}
\label{tab:milp_mmilp_app}
\end{table}

\begin{table}[]
\begin{tabular}{|l|l|l|l|l|c|l|l|l|l|l|}
\hline
\multicolumn{8}{|l|}{{\color[HTML]{000000} \textbf{Benchmark Information}}}                                                                                                                                                                                                                                               & {\color[HTML]{000000} \textbf{UA}}       & {\color[HTML]{000000} \textbf{BDA}}      & {\color[HTML]{000000} \textbf{MaxMILP}}  \\ \hline
{\color[HTML]{000000} \textbf{ID}} & {\color[HTML]{000000} \textbf{Name}}       & {\color[HTML]{000000} \textbf{$|\boldsymbol{\classifiers|}$}} & {\color[HTML]{000000} \textbf{$|\boldsymbol{\attackSet|}$}} & {\color[HTML]{000000} \textbf{$|\boldsymbol{\datasetLabelled|}$}} & {\color[HTML]{000000} \textbf{dim}} & {\color[HTML]{000000} \textbf{Epsilon}} & {\color[HTML]{000000} \textbf{Alpha}} & {\color[HTML]{000000} \textbf{$\boldsymbol{\Val(\attackSet,\probDistr)}$}} & {\color[HTML]{000000} \textbf{$\boldsymbol{\Val(\attackSet,\probDistr)}$}} & {\color[HTML]{000000} \textbf{$\boldsymbol{\Val(\attackSet,\probDistr)}$}} \\ \hline
1           & mnist\_0\_1         & 3            & 2            & 4            & 7x7           & 0.1            & 1000             & 0.33              & 0.5               & 1*                \\ \hline
2           & mnist\_0\_1         & 3            & 2            & 4            & 7x7           & 0.2            & 100              & 0.33              & 0.5               & 1*                \\ \hline
3           & mnist\_0\_1\_2convs & 3            & 2            & 4            & 8x8           & 0.2            & 1000             & 0.33              & 0.5               & 1*                \\ \hline
4           & mnist\_0\_1\_2convs & 3            & 2            & 4            & 8x8           & 0.4            & 100              & 0.33              & 0.25              & 1*                \\ \hline
5           & mnist\_0\_1\_2convs & 3            & 2            & 4            & 8x8           & 0.9            & 100              & 0.33              & 0.25              & 1*                \\ \hline
6           & mnist\_0\_1\_2convs & 3            & 3            & 4            & 8x8           & 0.9            & 100              & 0.33              & 0.25              & 1*                \\ \hline
7           & mnist\_0\_1         & 3            & 2            & 4            & 8x8           & 0.4            & 1000             & 0.33              & 0.5               & 1*                \\ \hline
8           & mnist\_0\_1         & 3            & 2            & 4            & 8x8           & 0.9            & 1000             & 0.33              & 0.5               & 1*                \\ \hline
9           & mnist\_0\_1         & 3            & 3            & 4            & 8x8           & 0.9            & 60               & 0.33              & 0.5               & 1*                \\ \hline
10          & mnist\_0\_1         & 3            & 4            & 4            & 8x8           & 0.9            & 60               & 0.33              & 0.5               & 1*                \\ \hline
11          & mnist\_4\_5         & 3            & 2            & 4            & 10x10         & 0.2            & 100              & 0.33              & 0.25              & 1*                \\ \hline
12          & mnist\_4\_5         & 3            & 4            & 4            & 10x10         & 0.3            & 50               & 0.33              & 0.75              & 1                 \\ \hline
13          & mnist\_4\_5         & 3            & 4            & 4            & 10x10         & 0.9            & 50               & 0.33              & 0.5               & 1                 \\ \hline
14          & mnist\_7\_8         & 3            & 4            & 4            & 6x6           & 0.1            & 60               & 0.33              & 0.5               & 1                 \\ \hline
15          & mnist\_7\_8         & 3            & 10           & 4            & 6x6           & 0.1            & 60               & 0.33              & 0.5               & 1                 \\ \hline
16          & mnist\_4\_5         & 3            & 2            & 4            & 10x10         & 0.1            & 1000             & 0.33              & 0.75              & 1*                \\ \hline
17          & mnist\_4\_5         & 3            & 3            & 4            & 10x10         & 0.1            & 80               & 0.33              & 0.5               & 1                 \\ \hline
18          & mnist\_4\_5         & 3            & 2            & 4            & 10x10         & 0.15           & 80               & 0.33              & 0.5               & 1*                \\ \hline
19          & mnist\_4\_5         & 3            & 2            & 4            & 10x10         & 0.2            & 100              & 0.33              & 0.5               & 1*                \\ \hline
\end{tabular}
\caption{Attacker Comparison}
\label{tab:attacker_comparison_app}
\end{table}

\end{document}